\documentclass{article}
\usepackage{spconf,amsmath,graphicx,hyperref}

\usepackage{algorithm}
\usepackage{algorithmic}
\usepackage{amssymb}
\usepackage{amsfonts}       
\usepackage{booktabs}
\usepackage{multirow}
\DeclareMathAlphabet{\mathbbold}{U}{bbold}{m}{n}

\newcommand{\eg}{\textit{e.g.}}
\newcommand{\myname}{QL-Adapter}
\newcommand{\mone}{ILFM}
\newcommand{\mtwo}{CMAM}
\newcommand{\dataset}{QL-Dataset}

\title{Instruction Guided Multi Object Image Editing with Quantity and Layout Consistency}

\name{Jiaqi Tan, Fangyu Li, Yang Liu$^{^\dagger}$\thanks{$^\dagger$ Corresponding author. Email: yang.liu@bupt.edu.cn}}
\address{Beijing University of Posts and Telecommunications \\ School of Digital Media and Design Art}
\begin{document}
\ninept
\maketitle
\begin{abstract}
Instruction driven image editing with standard CLIP text encoders often fails in complex scenes with many objects. We present \myname, a framework for multiple object editing that tackles two challenges: enforcing object counts and spatial layouts, and accommodating diverse categories. \myname\ consists of two core modules: the Image-Layout Fusion Module (\mone) and the Cross-Modal Augmentation Module (\mtwo). \mone\ fuses layout priors with ViT patch tokens from the CLIP image encoder to strengthen spatial structure understanding. \mtwo\ injects image features into the text branch to enrich textual embeddings and improve instruction following. We further build \textbf{\dataset}, a benchmark that spans broad category, layout, and count variations, and define the task of quantity and layout consistent image editing (QL-Edit). Extensive experiments show that \myname\ achieves state of the art performance on QL-Edit and significantly outperforms existing models.

\end{abstract}
\section{Introduction}
\label{sec:intro}


In recent years, diffusion models~\cite{nichol2021glide,rombach2022high,saharia2022photorealistic} have achieved remarkable progress in image generation and editing. Controllable editing methods can be broadly divided into two paradigms: text driven editing and mask based editing. Text driven approaches~\cite{hertz2022prompt,brooks2023instructpix2pix,shen2025long,huang2024smartedit} rely on natural language instructions without explicit spatial constraints, aiming to strengthen language understanding for instruction based editing. InstructPix2Pix~\cite{brooks2023instructpix2pix} trains a conditional diffusion model on synthetic image--instruction pairs generated by GPT\mbox{-}3~\cite{floridi2020gpt} and Stable Diffusion, enabling effective instruction following. SmartEdit~\cite{huang2024smartedit} replaces CLIP with a vision language model to enhance text comprehension. InstantStyle~\cite{wang2024instantstyle} observes that the fourth and sixth blocks in diffusion pipelines correspond to layout and style, respectively.
Mask based approaches~\cite{zhang2025context,shen2025imagdressing} provide stronger spatial control by constraining edits to user defined or automatically generated regions. LEDITS~\cite{tsaban2023ledits} combines DDPM inversion with semantic guidance for flexible editing without altering the architecture. IC\mbox{-}Edit~\cite{zhang2025context} integrates LoRA\mbox{-}MoE fine tuning with VLM guided masked sampling to achieve high quality zero shot editing with low data and parameter costs.

Recent work strengthens layout control to improve alignment. Inpaint Anything~\cite{yu2023inpaint} leverages SAM~\cite{kirillov2023segment} to introduce semantic masks for flexible spatial control. Other studies explicitly inject layout information to reinforce structural and semantic consistency~\cite{zhao2019image,shen2025imaggarment,shen2024imagpose}. IMAGHarmony~\cite{shen2025imagharmony}, built upon IP\mbox{-}Adapter~\cite{ye2023ip}, uses auxiliary text to tighten alignment between editing conditions and image features, and further defines quantity\mbox{ }and layout consistent image editing (QL-Edit).
Despite these advances, existing methods often struggle in multi object scenarios, where controlling object count and spatial layout remains challenging. Prompt driven methods exhibit weak counting ability (\eg, generating four or seven cats instead of exactly six), while mask based methods may confuse ``multiple independent objects'' with ``parts of a single object,'' leading to counting errors, layout disorder, and style inconsistency. A primary cause is the misalignment between editing conditions and targets in both structural and semantic modeling.

To address these challenges, we propose \myname, a framework that aligns textual control conditions with image structure while imposing explicit quantity and layout constraints for stronger controllability. Specifically, \myname\ contains two modules: \textcircled{1} \textbf{Image-Layout Fusion Module} (\mone), which fuses layout information with visual features to model object counts and capture hierarchical spatial relations; and \textcircled{2} \textbf{Cross-Modal Augmentation Module} (\mtwo), which enables bidirectional cross modal interaction and feeds augmented representations into cross attention to improve text--image alignment. The \myname~pipeline takes an image and auxiliary text as input, extracts layout maps and ViT patch features, and adapts IP-Adapter to edit a single image.
Moreover, we introduce \dataset, a benchmark for multi object editing consistency that spans diverse settings of object count, category, and layout, with both structural and semantic evaluation metrics. Extensive experiments show that \myname~outperforms existing methods by a significant margin in object count accuracy, layout fidelity, and overall generation quality.

\begin{figure*}[t!]
\centering
\includegraphics[width=1.0\textwidth]{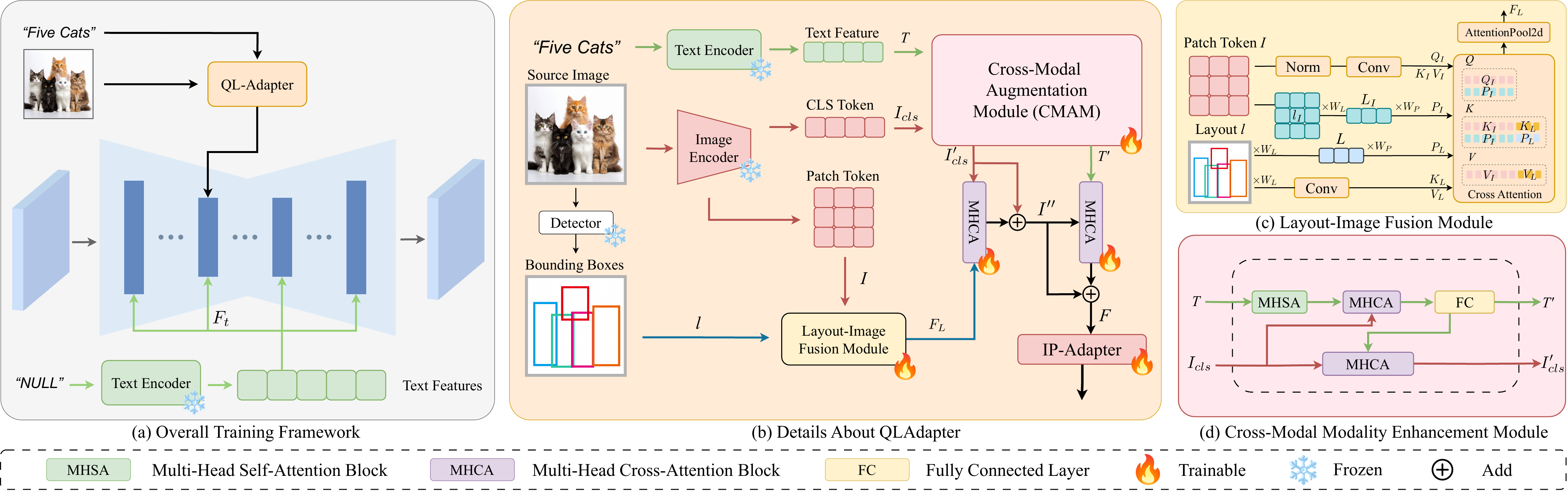}
\vspace{-15pt}
\caption{Overview of \myname. (a) Training and inference pipeline. During training, textual and visual features are injected into the fourth down block together with empty text. At inference, the empty text is replaced by an edit prompt. (b) Internal architecture with the Image and Layout Fusion Module (\mone) and the Cross Modal Augmentation Module (\mtwo) to model quantity and layout. (c) and (d) Detailed designs of \mone\ and \mtwo, respectively.}
\label{fig:QLAdapter}
\vspace{-10pt}
\end{figure*}

\section{Method}
\label{sec:method}

Our framework (Fig.~\ref{fig:QLAdapter}) enhances count and layout awareness in diffusion models by injecting auxiliary textual and layout cues. It contains two modules: Image Layout Fusion (\mone) and Cross Modal Augmentation (\mtwo). \mone\ encodes explicit bounding box layouts instead of masks and fuses them with ViT patch tokens from the CLIP image encoder to strengthen spatial structure modeling. \mtwo\ enables bidirectional interaction between visual and auxiliary textual features, making visual tokens responsive to text and enriching text embeddings with complementary visual semantics.

\subsection{Image-Layout Fusion Module}
\label{sec:\mone}
The detailed structure of the Image-Layout Fusion Module is shown in Fig.~\ref{fig:QLAdapter}.(c). We first extract the layout information using a forzen detector (YOLOv11~\cite{khanam2024yolov11}). A layout $l = \{l_1, l_2, \ldots, l_n\}$ is a collection of information about bounding boxes, where each $l_i$ represents the normalized bounding box coordinates $l_i = (x_0^i, y_0^i, x_1^i, y_1^i) \in [0,1]^4$. To support variable-length inputs, we set $l_0 = (0, 0, 1, 1)$ and define the total number of layouts as $n$. The empty object layout parts are padded with $(0, 0, 0, 0)$.
After the data processing operation, the layout embedding can be obtained through the matrix $W_L\in \mathbb{R}^{4\times d_L}$ as follows:
\vspace{-4pt}
\begin{equation}
    L = lW_L,
\label{eq:L}
\end{equation}
where $d_L$ is the dimension of the layout embedding.
Through the CLIP encoder, we can obtain the global representation (CLS token) and the local representation (patch tokens). To effectively integrate position and size information, we need to incorporate region concepts containing positional and size information to process the patch tokens. Specifically, for the patch tokens $I \in \mathbb{R}^{h \times w \times d_I}$, we need to align them with the bounding boxes layout on a patch-by-patch basis and obtain the image embedding by sharing the matrix weights \(W_L\) from Eq \ref{eq:L}. as follows:
\vspace{-8pt}
\begin{equation}
    l_{I_{u,v}}=(\frac{u}{h},\frac{v}{w},\frac{u+1}{h},\frac{v+1}{w}),\quad L_I=l_IW_L,
\label{eq:l_iuv}
\end{equation}
where $I_{u,v}$ represents the $u^{th}$ row and $v^{th}$ column patch of $I$, $u \in [0,h),v\in [0,w)$. With each patch has a corresponding bounding box, the spatial positions of the image and the layout are unified.




During the cross-attention stage, the image features $I$ and the layout features $L$ are first processed separately. Specifically, the image features are normalized and convolved to obtain $Q_I, K_I, V_I$ such that $Q_I,K_I,V_I = \text{Conv}(\text{Norm}(I))$, while the layout features are convolved to produce $K_L, V_L$ as $K_L,V_L = \text{Conv}(L)$. In parallel, a projection matrix $W_P \in \mathbb{R}^{d_L \times d_I}$ is employed to transform the layout embedding $L$ into positional embeddings, yielding $P_I=L_IW_P$ and $P_L=LW_P$. Based on these representations, the inputs to the cross-attention module are constructed as follows: the query is defined as $Q = Q_I \oplus_c P_I$, the key as $K = (K_I \oplus_c K_L)\oplus_s(P_I \oplus_c P_L)$, and the value as $V = V_I \oplus_s V_L$, where $\oplus_c$ and $\oplus_s$ denote concatenation along the channel dimension and the sequence length dimension, respectively. 
These inputs are then fed into the multi-head attention mechanism, formulated as follows:
\vspace{-4pt}
\begin{equation}
    \text{Attention}(Q,K,V)=\text{Softmax}(\tfrac{QK^T}{\sqrt{d_k}})V,
    \label{eq:attention}
\end{equation} 
finally, the attention outputs are aggregated through AttentionPool2d~\cite{radford2021learning}, thereby producing the feature representation $F_L$ with the same dimensionality as the CLS token.

\subsection{Cross-Modal Augmentation Module}
The detailed structure of the Cross-Modal Augmentation Module is shown in Fig.~\ref{fig:QLAdapter}.(d). To better extract intra-modal features and fuse cross-modal features, we define Multi-Head Self-Attention as Eq.~\ref{eq:attention} and Multi-Head Cross-Attention as follows:
\vspace{-6pt}
\begin{align}
    \text{MHCA}(f_1,f_2)&= \text{Softmax}\left(\frac{Q_1K_2^T}{\sqrt{d_k}}\right)V_2 \label{eq:MHCA},
\end{align}
where \(Q_1 = f_1 W_Q\) represents the Query extracted from \( f_1 \), \(K_2 = f_2 W_K\) and \(V_2 = f_2 W_V\) represents the Key and Value extracted from \( f_2 \). In \mtwo, we perform cross-modal interaction between the text encoding $T$ and the image CLS encoding $I_{\text{cls}}$, and return $T'$ and $I_{\text{cls}}'$. Specifically, we first pass the text information $T$ through the MHSA module. Then, $T$ is used as the Query, and $I_{\text{cls}}$ is used as both the Key and Value for interaction through the MHCA. Finally, we feed the output into a fully connected layer to obtain the processed text information $T'$. After that, we use the image feature $I_{\text{cls}}$ as the Query and $T'$ as both the Key and Value for information interaction, obtaining the output image feature $I_{\text{cls}}'$. 


\subsection{\myname}
The detailed structure of the \myname~is shown in Fig.~\ref{fig:QLAdapter}.(b). In the \myname, we fuse the layout features $L$ obtained from \mone, the enhanced modal features $T'$ and $I_{cls}'$ obtained from \mtwo, and finally output the fused feature $F$ with the same dimensionality as the visual features, thus maintaining consistency with the weights of the pre-trained IP-Adapter. Specifically, we first use $I_{cls}'$ as the Query and $L$ as both the Key and Value for fusion through MHCA in Eq.~\ref{eq:MHCA} and a residual structure, in order to obtain the layout-enhanced image features. The equation is as follows:
\vspace{-3pt}
\begin{equation}
    I''=I_{cls}' + \text{MHCA}(I_{cls}',F_L),
\end{equation}
then, we fuse the obtained feature $I''$ with the text-enhanced features to gain sensitivity to the text. The equation is as follows:
\vspace{-3pt}
\begin{equation}
    F=I'' + \text{MHCA}(I'',T'),
\end{equation}
finally, we obtain the fused feature $F$ with position and quantity awareness, which has the same dimension as $I_{\text{cls}}$. We then feed it into the IP attention~\cite{ye2023ip} for subsequent training and fine-tuning operations to achieve control over the image.

\subsection{Training and Inference}
We inject the mixed feature $F$ into the diffusion model using a dual-branch IP-Adapter~\cite{ye2023ip}. One branch receives an empty text token and uses frozen cross-attention to avoid textual interference. The other branch inputs $F$ into a trainable IP-attention module inserted at the fourth down block, which is most sensitive to layout~\cite{wang2024instantstyle}. The attention fusion is
\vspace{-6pt}
\begin{equation}
Z_{\text{new}}=\text{Softmax}\!\left(\frac{QK_t^{\top}}{\sqrt{d}}\right)V_t \;+\; \lambda\,\text{Softmax}\!\left(\frac{QK_f^{\top}}{\sqrt{d}}\right)V_f,
\label{eq:z}
\end{equation}
where $Q=Z_tW_q$ is projected from the noisy latent $Z_t$, $K_t=F_tW_k^{t}$ and $V_t=F_tW_v^{t}$ come from the empty text encoding $F_t$, and $K_f=FW_k^{f}$ and $V_f=FW_v^{f}$ come from the mixed feature $F$. The scalar $\lambda$ controls the contribution of the quantity and layout cues.

During training, we optimize only the IP-attention at the fourth down block and keep the SDXL backbone frozen. The objective is
\begin{equation}
\mathcal{L}=\mathbb{E}_{x_0,\epsilon,F_t,F,t}\bigl\|\epsilon-\epsilon_{\theta}(x_t,F_t,F,t)\bigr\|^2,
\end{equation}
where $x_0$ is the ground-truth image, $\epsilon$ is Gaussian noise, $x_t$ is the noisy image at timestep $t$, and $\epsilon_{\theta}$ is the predicted noise.

At inference, the empty text branch is replaced by an edit prompt. We apply classifier-free image guidance by randomly dropping the image condition:
\vspace{-6pt}
\begin{equation}
\hat{\epsilon}_{\theta}(x_t,F_t,F,t)=w\,\epsilon_{\theta}(x_t,F_t,F,t)+(1-w)\,\epsilon_{\theta}(x_t,t),
\end{equation}
where $w$ controls the guidance strength.

\begin{figure}[t]
\centering
\includegraphics[width=0.48\textwidth]{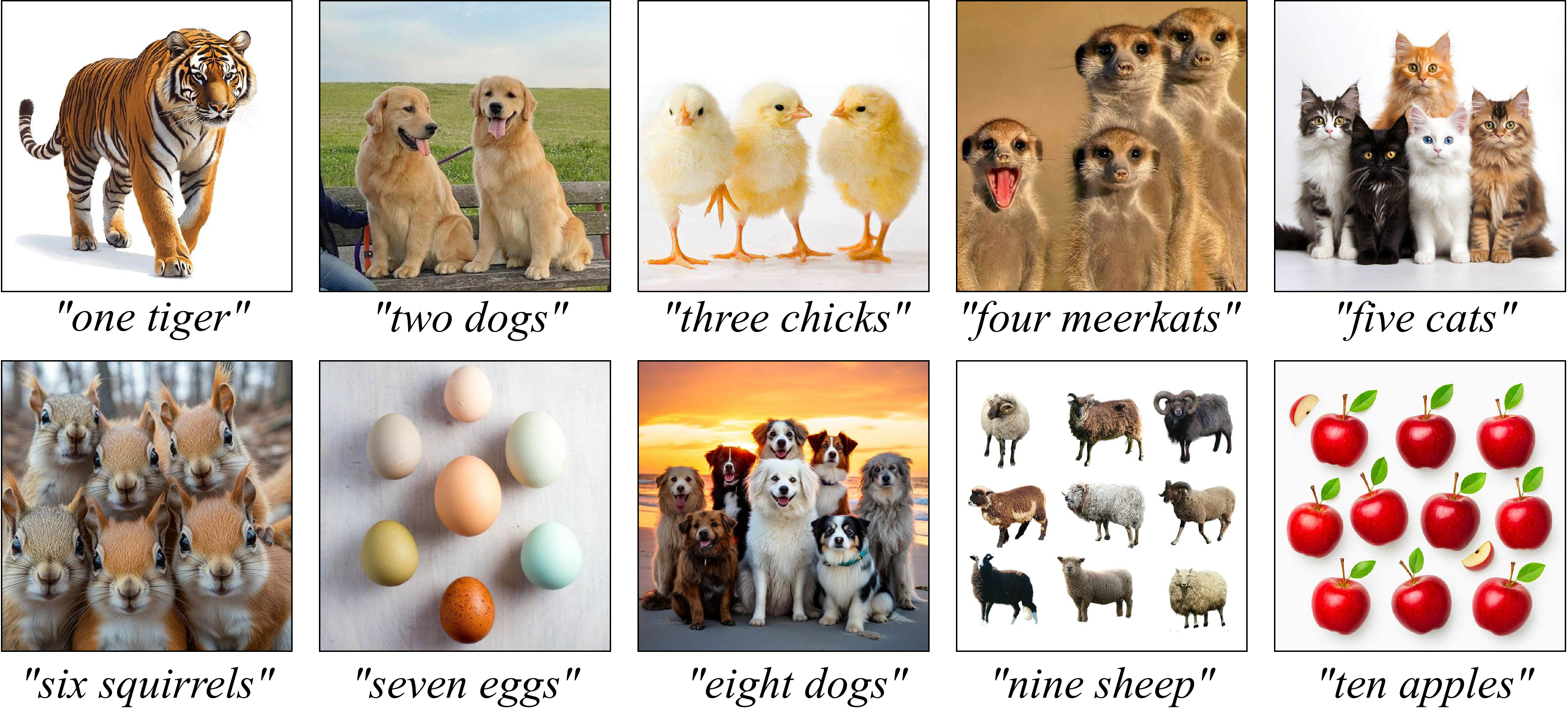}
\vspace{-10pt}
\caption{Examples from \dataset. Each image–text pair specifies a quantity between 1 and 10. For every quantity, 10 category diverse images are included, yielding 100 images in total.}
\label{fig:QLDataset}
\end{figure}

\section{Experiments}
\subsection{Implementation Details}

\noindent\textbf{Dataset.}
To evaluate whether image editing models preserve object quantity, we introduce \dataset, a benchmark that targets quantity aware evaluation missing from existing datasets. \dataset~contains 100 image and caption pairs with object counts from 1 to 10, with 10 diverse examples per count level, as shown in Fig.~\ref{fig:QLDataset}. Captions are written in a simple style (\eg, ``six dogs'') to reduce ambiguity in number expression. Images are collected from COCO, copyright free images on Pinterest, and synthetic data produced by SDXL, covering both real world and synthetic domains with diverse scenes and arrangements. For each reference image we design five quantity editing instructions, including adding or removing a specified number of objects, yielding 500 test instructions that support a comprehensive assessment of counting fidelity under varied edits.

\noindent\textbf{Evaluation Metrics.}
We report three quantitative metrics to measure number and layout awareness. Object Accuracy (OA)~\cite{liu2019point} counts a detection $\hat{g}$ as correct if its Intersection over Union with the ground truth box $g$ exceeds $0.5$, and the number of detected boxes matches the ground truth count:
\vspace{-6pt}
\begin{equation}
\operatorname{IoU}(g,\hat{g})=\frac{\mathrm{area}(g\cap \hat{g})}{\mathrm{area}(g\cup \hat{g})}.
\label{eq:iou}
\end{equation}
Average Precision (AP)~\cite{cho2024diagnostic} is the area under the Precision Recall curve:
\vspace{-6pt}
\begin{equation}
\mathrm{AP}=\int_{0}^{1} P(R)\, dR .
\label{eq:ap}
\end{equation}
Here $P(R)$ is computed from Precision and Recall:
\vspace{-6pt}
\begin{equation}
    \mathrm{Precision}=\frac{TP}{TP+FP},\ \mathrm{Recall}=\frac{TP}{TP+FN}.
\end{equation}
Image Reward (IR)~\cite{xu2023imagereward} is a human preference based score that, given a text prompt $x$ and an image $I$, outputs $R(I\mid x)$ learned from large scale human feedback.

\noindent\textbf{Hyperparameters.}
All experiments are conducted on a single NVIDIA A6000 GPU. The backbone is SDXL, and IP attention is initialized with the pretrained IP Adapter. Multihead cross attention (MHCA) and multihead self attention (MHSA) use eight heads and apply a \(5\%\) condition dropout to enable classifier free guidance (CFG). We use $512\times512$ resolution, AdamW with a learning rate of \(2.5\times10^{-4}\), and train for 2100 steps. At inference, we use 30 denoising steps, a CFG scale of 5, and set other attention scales to 1. Training a single image takes roughly 20 minutes.

\subsection{Comparison with State of the Art Methods}

\begin{figure}[t!]
\centering
\includegraphics[width=0.48\textwidth]{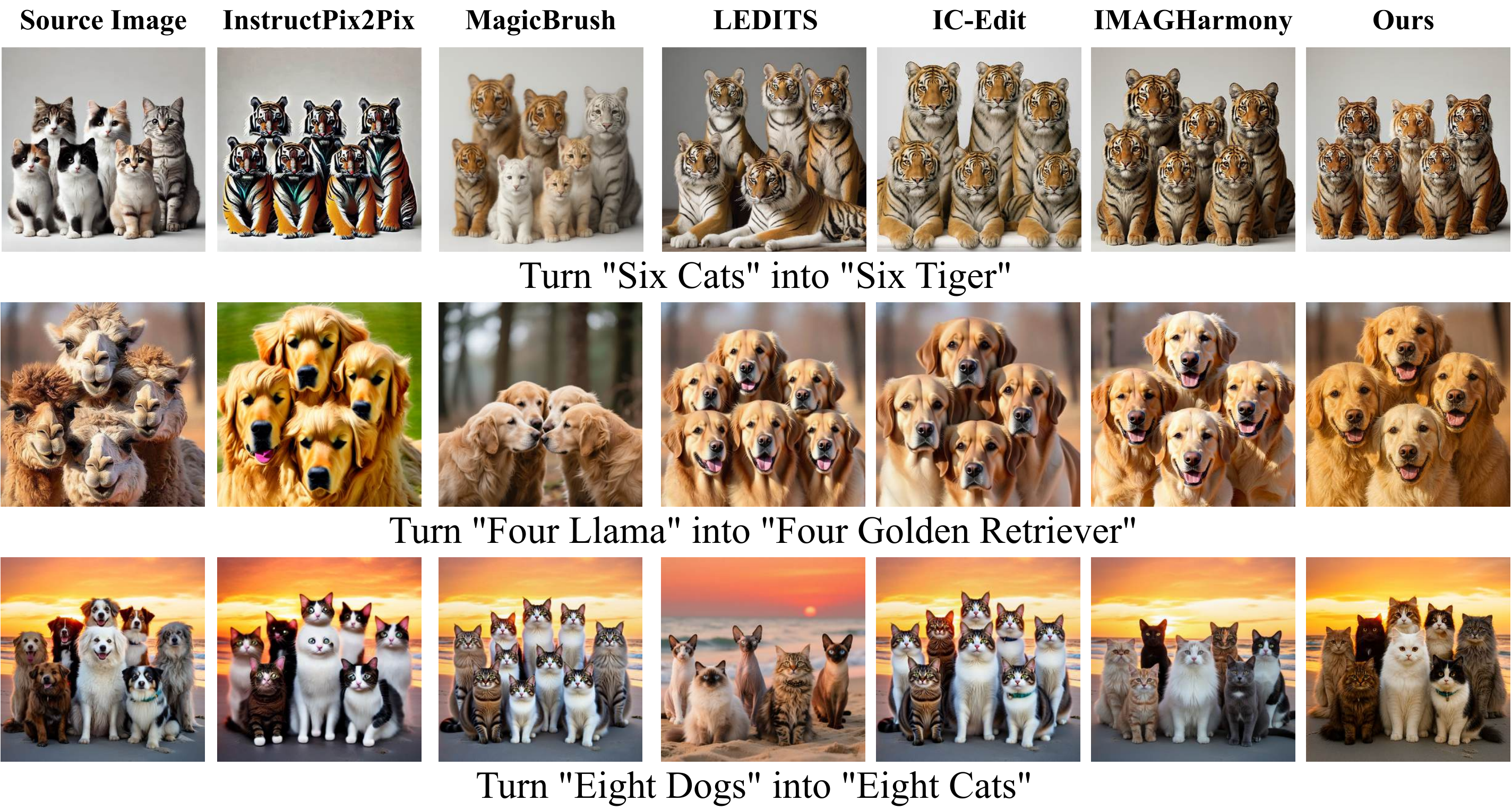}
\vspace{-15pt}
\caption{Qualitative Comparison Results with several SoTA models.}
\label{fig:Result}
\vspace{-15pt}
\end{figure}


\noindent\textbf{Qualitative Results.}
We compare \myname\ with SoTA text-driven editing methods InstructPix2Pix~\cite{brooks2023instructpix2pix} and MagicBrush~\cite{zhang2023magicbrush}, as well as mask-based editing methods LEDITS~\cite{tsaban2023ledits} and IC-Edit~\cite{zhang2025context}. The qualitative results are shown in Fig.~\ref{fig:Result}. InstructPix2Pix performs well when the number of target objects is small, but it suffers from object mixing when there are many targets (\eg, in the third-row instruction, the middle cat is incorrectly mixed). MagicBrush fails on the QL-Edit task, unable to generate images with the correct number and layout. LEDITS, due to the use of masks, can produce relatively good positional effects but performs poorly when dealing with fine-grained structures inside the masked areas. IC-Edit performs relatively well on the QL-Edit task but still shows deviations in layout. In contrast, our model effectively preserves consistency in layout, quantity, and background during image editing. Compared with the original image, it not only generates the correct number of objects but also maintains a high degree of similarity in layout.

\begin{table}[t]
\centering
\resizebox{\linewidth}{!}{
\begin{tabular}{l|ccc|ccc}
\hline
\multirow{2}{*}{\textbf{Method}} & \multicolumn{3}{c|}{\textbf{Class Editing ($\uparrow$)}} & \multicolumn{3}{c}{\textbf{Scene Editing ($\uparrow$)}} \\ \cline{2-7}
 & OA & AP & IR & OA & AP & IR \\ \hline
InstructPix2Pix~\cite{brooks2023instructpix2pix} & 64.8 & 63.5 & 0.355 & 90.0 & 89.3 & 0.081 \\
MagicBrush~\cite{zhang2023magicbrush} & 68.2 & 67.0 & 0.421 & 89.2 & 88.5 & 0.097 \\
LEDITS~\cite{tsaban2023ledits} & 44.2 & 41.0 & 0.449 & 75.3 & 70.1 & -0.142 \\
IC-Edit~\cite{zhang2025context} & 71.6 & 72.2 & 0.529 & 92.1 & 89.0 & 0.183 \\
IMAGHarmony~\cite{shen2025imagharmony} & 73.4 & 70.8 & 0.574 & 91.4 & 87.8 & 0.203 \\ \hline
\textbf{\myname} & \textbf{76.2} & \textbf{73.7} & \textbf{0.611} & \textbf{92.3} & \textbf{90.3} & \textbf{0.206} \\ \hline
\end{tabular}}
\vspace{-6pt}
\caption{Quantitative Comparison Results of different methods on Class Editing and Scene Editing tasks.}
\label{tab:comparison}
\vspace{-10pt}
\end{table}


\noindent\textbf{Quantitative Results.}
The quantitative results are shown in Table~\ref{tab:comparison}, where our proposed \myname\ consistently outperforms existing methods across both Class Editing and Scene Editing tasks. In terms of object accuracy (OA), \myname\ achieves the highest scores of 76.2 and 92.3, surpassing the second-best method IC-Edit by 4.7 and 0.2 points, respectively. This demonstrates its superior capability in precise object number control for both foreground class editing and background scene editing. For average precision (AP), which measures layout consistency with reference annotations, \myname\ achieves 73.7 and 90.3, again outperforming all baselines. The improvements over IC-Edit (72.1 and 88.9) highlight that \myname\ not only generates the correct number of objects but also ensures accurate spatial alignment with the target layout. In addition to quantity and layout control, \myname\ also achieves the best performance in Image Reward (IR), scoring 0.611 and 0.206 for Class Editing and Scene Editing, respectively. This indicates that \myname\ not only meets structural constraints but also generates images with higher visual quality and better instruction alignment compared to all baselines.


\subsection{Ablation Study}

\begin{figure}[t!]
\centering
\includegraphics[width=0.48\textwidth]{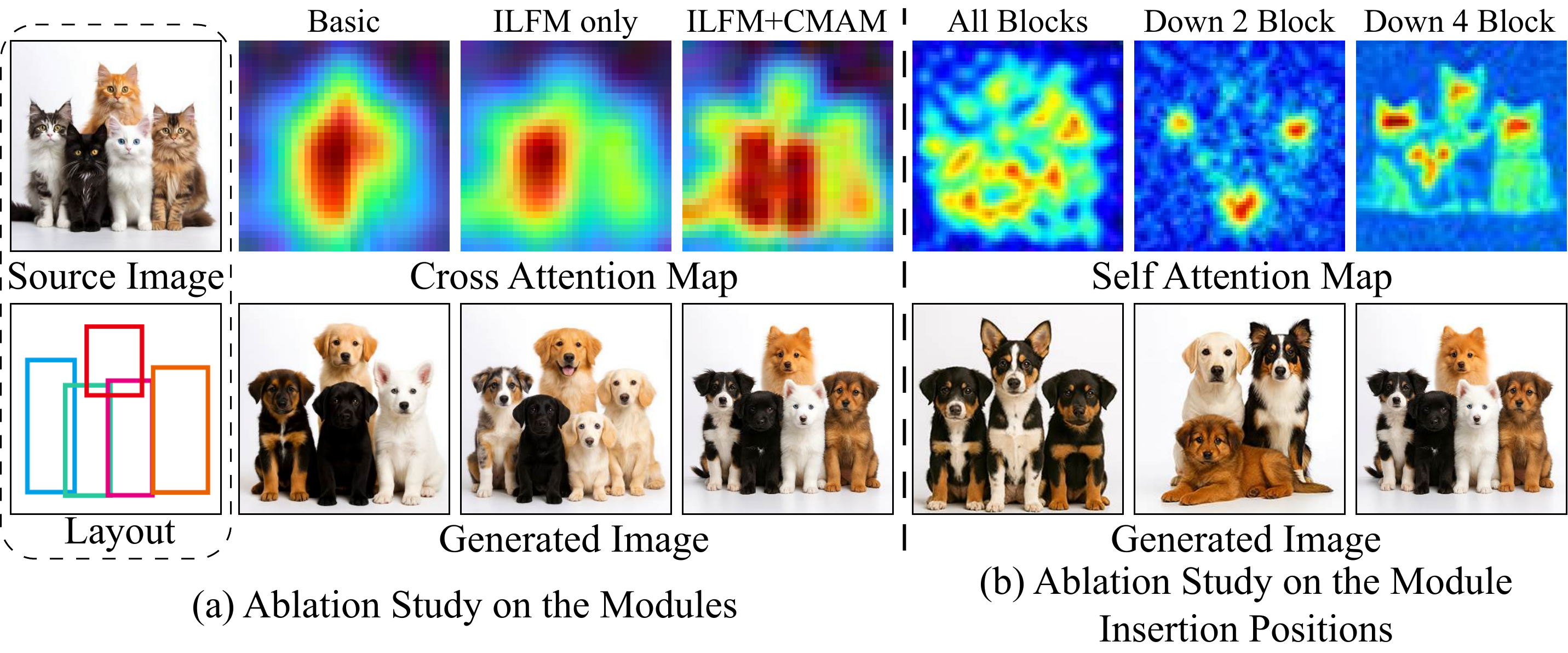}
\vspace{-15pt}
\caption{(a) Cross-attention maps at the fourth down block with outputs for different module combinations. (b) Influence of insertion positions on self-attention maps in the mid block and the corresponding generations.}
\label{fig:Ablation}
\vspace{-10pt}
\end{figure}

\noindent\textbf{Effectiveness of \mone.}
We visualized the Cross Attention layer where the \myname\ is inserted at the “Down 4 Block” location (see Fig.~\ref{fig:Ablation} (a)). Under the Basic setting, attention is mainly driven by image saliency, forming a single, centrally concentrated hotspot. This indicates that the model tends to focus on visually prominent regions while neglecting semantic and layout cues, which leads to limited spatial coverage and weak coupling between “where” and “what”, often causing errors in object count and placement. After introducing the Image-Layout Fusion Module (\mone), the attention heatmap shifts from a single peak to multiple peaks with broader spatial coverage, and the highlighted regions become aligned with the bounding boxes specified in the layout. This shows that \mone\ effectively injects layout priors into image feature computation, mitigating center bias, distributing attention more evenly across spatial positions, and forming coarse-grained spatial partitions that guide instance-level alignment.

\noindent\textbf{Effectiveness of \mtwo.}
Building upon \mone, the integration of the Cross-Modal Augmentation Module (\mtwo) further refines the attention distribution. As shown in Fig.~\ref{fig:Ablation} (a), the broad coverage brought by \mone\ is sharpened into multiple fine-grained hotspots: within each layout region, peaks become more distinct while energy leakage across regions is reduced. This indicates that \mtwo\ enhances semantic details and local discriminability, enabling the model to focus on more precise subregions. The second-row results demonstrate that this refined attention translates into improved image quality, validating that \mtwo, when combined with \mone, contributes to stronger instance-level precision and visual fidelity.


\noindent\textbf{Influence of Insertion Positions.}
The ablation on insertion positions, shown in Fig.~\ref{fig:Ablation} (b), reveals that placing the conditional module at the “Down 4 Block” achieves the best results. Specifically, the self-attention maps at the mid block show that this position yields more focused and semantically meaningful attention regions. By contrast, inserting at “Down 2 Block” provides only limited improvement due to insufficient global context, while inserting at “All Blocks” introduces excessive conditioning noise, leading to scattered attention. The generated images under the Down 4 Block setting exhibit sharper structures and higher visual fidelity, confirming that conditioning at deeper layers effectively balances global semantics and local details, enabling the diffusion model to produce more realistic and coherent outputs.



\begin{figure}[t!]
\centering
\includegraphics[width=0.48\textwidth]{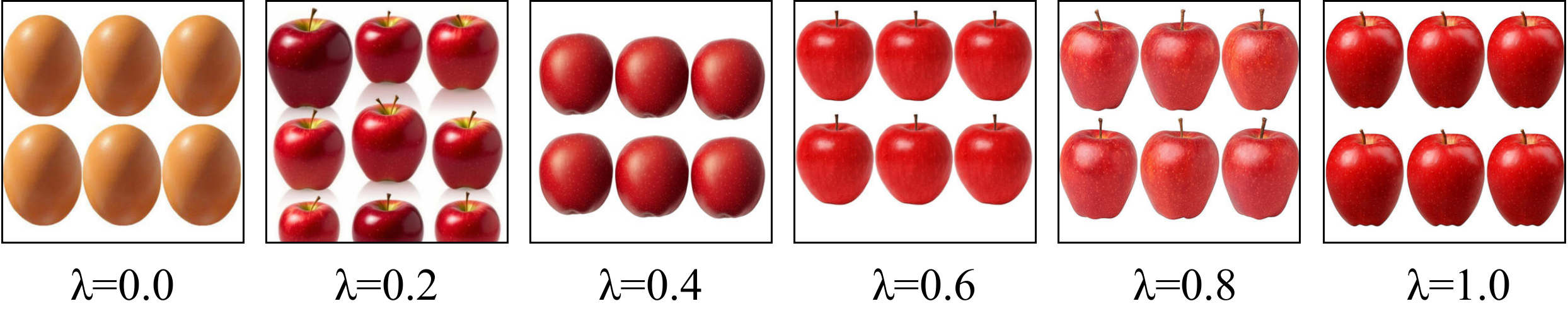}
\vspace{-15pt}
\caption{Visualization of the hyperparameter \(\lambda\).}
\label{fig:Ablation-lambda}
\vspace{-15pt}
\end{figure}

\noindent\textbf{Hyperparameter Ablation Study} To investigate the impact of the hyperparameter \(\lambda\) in Eq.~\ref{eq:z} on the image translation performance, we gradually increased \(\lambda\) from 0.0 to 1.0 to transform source images (eggs) into target images (red apples). As shown in Fig.~\ref{fig:Ablation-lambda}, when \(\lambda=0.0\), the generated image completely retains the source characteristics with no apple-like features. As \(\lambda\) increases, apple-related features, such as color and shape, gradually emerge, and the visual quality steadily improves. The optimal visual realism is achieved at \(\lambda=0.8\), where the generated apples exhibit natural shapes, vivid colors, and minimal artifacts. However, when \(\lambda\) is further increased to 1.0, the generated apples become over-smoothed and start to regain unnatural egg-like textures. This demonstrates that an intermediate \(\lambda\) (\eg, 0.8) provides the best trade-off between preserving target characteristics and avoiding visual distortions; therefore, all subsequent experiments are conducted with \(\lambda=0.8\).

\section{Conclusion}
\label{sec:conclusion}

We presented \myname, a framework for quantity and layout consistent image editing with diffusion models. \myname\ integrates the Image Layout Fusion Module (\mone) to encode spatial structure and the Cross Modal Augmentation Module (\mtwo) to strengthen text to image alignment. This design addresses the key challenges of controlling object count, preserving layout fidelity, and following editing instructions.
Experiments on \dataset\ show that \myname\ surpasses text driven and mask based baselines in object count accuracy, spatial alignment, and perceptual quality. Ablation studies confirm the complementary roles of \mone\ and \mtwo, enabling precise and flexible multi object editing. We believe \myname\ offers a principled foundation for image editing and opens directions for fine grained, instruction driven visual generation.

\vfill\pagebreak




{\small
\bibliographystyle{IEEEbib}
\bibliography{strings,refs}

\begin{thebibliography}{10}

\bibitem{nichol2021glide}
Alex Nichol, Prafulla Dhariwal, Aditya Ramesh, Pranav Shyam, Pamela Mishkin, Bob McGrew, Ilya Sutskever, and Mark Chen,
\newblock ``Glide: Towards photorealistic image generation and editing with text-guided diffusion models,''
\newblock {\em arXiv preprint arXiv:2112.10741}, 2021.

\bibitem{rombach2022high}
Robin Rombach, Andreas Blattmann, Dominik Lorenz, Patrick Esser, and Bj{\"o}rn Ommer,
\newblock ``High-resolution image synthesis with latent diffusion models,''
\newblock in {\em Proceedings of the IEEE/CVF conference on computer vision and pattern recognition}, 2022, pp. 10684--10695.

\bibitem{saharia2022photorealistic}
Chitwan Saharia, William Chan, Saurabh Saxena, Lala Li, Jay Whang, Emily~L Denton, Kamyar Ghasemipour, Raphael Gontijo~Lopes, Burcu Karagol~Ayan, Tim Salimans, et~al.,
\newblock ``Photorealistic text-to-image diffusion models with deep language understanding,''
\newblock {\em Advances in neural information processing systems}, vol. 35, pp. 36479--36494, 2022.

\bibitem{hertz2022prompt}
Amir Hertz, Ron Mokady, Jay Tenenbaum, Kfir Aberman, Yael Pritch, and Daniel Cohen-Or,
\newblock ``Prompt-to-prompt image editing with cross attention control,''
\newblock {\em arXiv preprint arXiv:2208.01626}, 2022.

\bibitem{brooks2023instructpix2pix}
Tim Brooks, Aleksander Holynski, and Alexei~A Efros,
\newblock ``Instructpix2pix: Learning to follow image editing instructions,''
\newblock in {\em Proceedings of the IEEE/CVF conference on computer vision and pattern recognition}, 2023, pp. 18392--18402.

\bibitem{shen2025long}
Fei Shen, Cong Wang, Junyao Gao, Qin Guo, Jisheng Dang, Jinhui Tang, and Tat-Seng Chua,
\newblock ``Long-term talkingface generation via motion-prior conditional diffusion model,''
\newblock {\em arXiv preprint arXiv:2502.09533}, 2025.

\bibitem{huang2024smartedit}
Yuzhou Huang, Liangbin Xie, Xintao Wang, Ziyang Yuan, Xiaodong Cun, Yixiao Ge, Jiantao Zhou, Chao Dong, Rui Huang, Ruimao Zhang, et~al.,
\newblock ``Smartedit: Exploring complex instruction-based image editing with multimodal large language models,''
\newblock in {\em Proceedings of the IEEE/CVF Conference on Computer Vision and Pattern Recognition}, 2024, pp. 8362--8371.

\bibitem{floridi2020gpt}
Luciano Floridi and Massimo Chiriatti,
\newblock ``Gpt-3: Its nature, scope, limits, and consequences,''
\newblock {\em Minds and machines}, vol. 30, no. 4, pp. 681--694, 2020.

\bibitem{wang2024instantstyle}
Haofan Wang, Matteo Spinelli, Qixun Wang, Xu~Bai, Zekui Qin, and Anthony Chen,
\newblock ``Instantstyle: Free lunch towards style-preserving in text-to-image generation,''
\newblock {\em arXiv preprint arXiv:2404.02733}, 2024.

\bibitem{zhang2025context}
Zechuan Zhang, Ji~Xie, Yu~Lu, Zongxin Yang, and Yi~Yang,
\newblock ``In-context edit: Enabling instructional image editing with in-context generation in large scale diffusion transformer,''
\newblock {\em arXiv preprint arXiv:2504.20690}, 2025.

\bibitem{shen2025imagdressing}
Fei Shen, Xin Jiang, Xin He, Hu~Ye, Cong Wang, Xiaoyu Du, Zechao Li, and Jinhui Tang,
\newblock ``Imagdressing-v1: Customizable virtual dressing,''
\newblock in {\em Proceedings of the AAAI Conference on Artificial Intelligence}, 2025, vol.~39, pp. 6795--6804.

\bibitem{tsaban2023ledits}
Linoy Tsaban and Apolin{\'a}rio Passos,
\newblock ``Ledits: Real image editing with ddpm inversion and semantic guidance,''
\newblock {\em arXiv preprint arXiv:2307.00522}, 2023.

\bibitem{yu2023inpaint}
Tao Yu, Runseng Feng, Ruoyu Feng, Jinming Liu, Xin Jin, Wenjun Zeng, and Zhibo Chen,
\newblock ``Inpaint anything: Segment anything meets image inpainting,''
\newblock {\em arXiv preprint arXiv:2304.06790}, 2023.

\bibitem{kirillov2023segment}
Alexander Kirillov, Eric Mintun, Nikhila Ravi, Hanzi Mao, Chloe Rolland, Laura Gustafson, Tete Xiao, Spencer Whitehead, Alexander~C Berg, Wan-Yen Lo, et~al.,
\newblock ``Segment anything,''
\newblock in {\em Proceedings of the IEEE/CVF international conference on computer vision}, 2023, pp. 4015--4026.

\bibitem{zhao2019image}
Bo~Zhao, Lili Meng, Weidong Yin, and Leonid Sigal,
\newblock ``Image generation from layout,''
\newblock in {\em Proceedings of the IEEE/CVF conference on computer vision and pattern recognition}, 2019, pp. 8584--8593.

\bibitem{shen2025imaggarment}
Fei Shen, Jian Yu, Cong Wang, Xin Jiang, Xiaoyu Du, and Jinhui Tang,
\newblock ``Imaggarment-1: Fine-grained garment generation for controllable fashion design,''
\newblock {\em arXiv preprint arXiv:2504.13176}, 2025.

\bibitem{shen2024imagpose}
Fei Shen and Jinhui Tang,
\newblock ``Imagpose: A unified conditional framework for pose-guided person generation,''
\newblock {\em Advances in neural information processing systems}, vol. 37, pp. 6246--6266, 2024.

\bibitem{shen2025imagharmony}
Fei Shen, Xiaoyu Du, Yutong Gao, Jian Yu, Yushe Cao, Xing Lei, and Jinhui Tang,
\newblock ``Imagharmony: Controllable image editing with consistent object quantity and layout,''
\newblock {\em arXiv preprint arXiv:2506.01949}, 2025.

\bibitem{ye2023ip}
Hu~Ye, Jun Zhang, Sibo Liu, Xiao Han, and Wei Yang,
\newblock ``Ip-adapter: Text compatible image prompt adapter for text-to-image diffusion models,''
\newblock {\em arXiv preprint arXiv:2308.06721}, 2023.

\bibitem{khanam2024yolov11}
Rahima Khanam and Muhammad Hussain,
\newblock ``Yolov11: An overview of the key architectural enhancements,''
\newblock {\em arXiv preprint arXiv:2410.17725}, 2024.

\bibitem{radford2021learning}
Alec Radford, Jong~Wook Kim, Chris Hallacy, Aditya Ramesh, Gabriel Goh, Sandhini Agarwal, Girish Sastry, Amanda Askell, Pamela Mishkin, Jack Clark, et~al.,
\newblock ``Learning transferable visual models from natural language supervision,''
\newblock in {\em International conference on machine learning}. PmLR, 2021, pp. 8748--8763.

\bibitem{liu2019point}
Yuting Liu, Miaojing Shi, Qijun Zhao, and Xiaofang Wang,
\newblock ``Point in, box out: Beyond counting persons in crowds,''
\newblock in {\em Proceedings of the IEEE/CVF conference on computer vision and pattern recognition}, 2019, pp. 6469--6478.

\bibitem{cho2024diagnostic}
Jaemin Cho, Linjie Li, Zhengyuan Yang, Zhe Gan, Lijuan Wang, and Mohit Bansal,
\newblock ``Diagnostic benchmark and iterative inpainting for layout-guided image generation,''
\newblock in {\em Proceedings of the IEEE/CVF Conference on Computer Vision and Pattern Recognition}, 2024, pp. 5280--5289.

\bibitem{xu2023imagereward}
Jiazheng Xu, Xiao Liu, Yuchen Wu, Yuxuan Tong, Qinkai Li, Ming Ding, Jie Tang, and Yuxiao Dong,
\newblock ``Imagereward: Learning and evaluating human preferences for text-to-image generation,''
\newblock {\em Advances in Neural Information Processing Systems}, vol. 36, pp. 15903--15935, 2023.

\bibitem{zhang2023magicbrush}
Kai Zhang, Lingbo Mo, Wenhu Chen, Huan Sun, and Yu~Su,
\newblock ``Magicbrush: A manually annotated dataset for instruction-guided image editing,''
\newblock {\em Advances in Neural Information Processing Systems}, vol. 36, pp. 31428--31449, 2023.

\end{thebibliography}
}
\end{document}